\documentclass[preprint]{elsarticle}
\usepackage{graphicx}
\usepackage[utf8]{inputenc}
\usepackage{graphicx,color}
\usepackage{algorithmic}
\usepackage{algorithm}
\usepackage{amssymb}
\usepackage{amsmath}
\usepackage{acronym}
\usepackage[]{todonotes}
\usepackage{hyperref}

\newcommand{\dataSize}{n}
\newcommand{\testSize}{s}
\newcommand{\rankingSize}{k}
\newcommand{\pairwiseSize}{h}

\newcommand{\attSize}{m}
\newcommand{\rulesetSize}{s}

\begin{document}

\begin{frontmatter}

%opening
\title{Preference rules for label ranking: Mining patterns in multi-target relations}
\author[inesc,liacs]{Cl{\'a}udio Rebelo de S{\'a}\corref{mycorrespondingauthor}}
\cortext[mycorrespondingauthor]{Corresponding author}
\ead{claudio.r.sa@inesctec.pt}
\author[hash]{Paulo Azevedo}
\ead{pja@di.uminho.pt}
\author[inesc]{Carlos Soares}
\ead{csoares@fe.up.pt}
\author[dcc]{Al{\'i}pio M{\'a}rio Jorge}
\ead{amjorge@fc.up.pt}
\author[liacs]{Arno Knobbe}
\ead{a.j.knobbe@liacs.leidenuniv.nl}

\address[inesc]{INESCTEC, Porto, Portugal}
\address[liacs]{LIACS, Leiden, Netherlands}
\address[hash]{HasLab, INESC TEC, Departamento de Inform{\'a}tica, Universidade do Minho}
\address[dcc]{DCC - Faculdade de Ciencias, Universidade do Porto}

%\institute{LIACS Universiteit Leiden\\
%\and INESCTEC Porto, Porto, Portugal \\
%\and Faculdade de Engenharia, Universidade do Porto
%\and DCC - Faculdade de Ciencias, Universidade do Porto
%\and HasLab, INESC TEC, Departamento de Inform{\'a}tica, Universidade do Minho \\
%\email{c.f.de.sa@liacs.leidenuniv.nl, csoares@fe.up.pt, amjorge@fc.up.pt, pja@di.uminho.pt} }

%\maketitle

\begin{abstract}
In this paper we investigate two variants of association rules for preference data, Label Ranking Association Rules and Pairwise Association Rules.
Label Ranking Association Rules (LRAR) are the equivalent of Class Association Rules (CAR) for the Label Ranking task.
In CAR, the consequent is a single class, to which the example is expected to belong to.
In LRAR, the consequent is a ranking of the labels.
The generation of LRAR requires special support and confidence measures to assess the similarity of rankings.
In this work, we carry out a sensitivity analysis of these similarity-based measures.
We want to understand which datasets benefit more from such measures and which parameters have more influence in the accuracy of the model.
Furthermore, we propose an alternative type of rules, the Pairwise Association Rules (PAR), which are defined as association rules with a set of pairwise preferences in the consequent.
While PAR can be used both as descriptive and predictive models, they are essentially descriptive models.
Experimental results show the potential of both approaches.
\end{abstract}


\end{frontmatter}

%\linenumbers

\section{Introduction}

Label ranking is a topic in the machine learning literature~\cite{furnkranz+05,cheng2009icml,VembuG10} that studies the problem of learning a mapping from instances to rankings over a finite number of predefined labels.
One characteristic that clearly distinguishes label ranking problems from classification problems is the order relation between the labels.
While a classifier aims at finding the true class on a given unclassified example, the label ranker will focus on the relative preferences between a set of labels/classes.
These relations represent relevant information from a decision support perspective, with possible applications in various fields such as elections, dominance of certain species over the others, user preferences, etc.

Due to its intuitive representation, Association Rules~\cite{Agrawal1994} have become very popular in data mining and machine learning tasks (e.g.\ Mining rankings~\cite{HenzgenH14}, Classification~\cite{liu1998integrating} and even Label Ranking~\cite{rebelosa2011}, etc).
The adaptation of AR for label ranking, Label Ranking Association Rules (LRAR)~\cite{rebelosa2011}, are similar to their classification counterpart, Class Association Rules (CAR)~\cite{liu1998integrating}.
LRAR can be used for predictive or descriptive purposes.

LRAR are relations, like typical association rules, between an antecedent and a consequent ($A \rightarrow C$), defined by interest measures.
The distinction lies in the fact that the consequent is a complete ranking.
Because the degree of similarity between rankings can vary, it lead to several interesting challenges.
For instance, how to treat rankings that are very similar but not exactly equal.
To tackle this problem, similarity-based interest measures were defined to evaluate LRAR.
Such measures can be applied to existing rule generation methods~\cite{rebelosa2011} (e.g.\ APRIORI~\cite{Agrawal1994}).

One important issue for the use of LRAR is the threshold that determines what should and should not be considered sufficiently similar.
Here we present the results of sensitivity analysis study to show how LRAR behave in different scenarios, to understand the effect of this threshold better.
Whether there is a rule of thumb or this threshold is data-specific is the type of questions we investigate here.
Ultimately we also want to understand which parameters have more influence in the predictive accuracy of the method.

Another important issue is related to the large number of distinct rankings.
Despite the existence of many competitive approaches in Label Ranking, Decision trees~\cite{todorovski+02,cheng2009icml}, \emph{k}-Nearest Neighbor~\cite{brazdil+03,cheng2009icml} or LRAR \cite{rebelosa2011}, problems with a large number of distinct rankings can be hard to predict.
%This is because finding patterns with complete and long rankings can be very hard.\csimportant{circular argument}
%For example, it is unexpected to observe 10 athletes finishing in the exact same relative order in distinct marathons.
%On the other hand, it is reasonable to expect to find, at least, one specific athlete always finishing before one other athlete.
%\csimportant{\ldots which contradicts the previous statement, as this would make the identification of patterns easier}
One real-world example with a relatively large number of rankings, is the sushi dataset~\cite{kamishima03}.
This dataset compares demographics of 5000 Japanese citizens with their preferred sushi types.
With only 10 labels, it has more than 4900 distinct rankings.
Even though it has been known in the preference learning community for a while, no results with high predictive accuracy have been published, to the best of our knowledge.
Cases like this have motivated the appearance of new approaches, e.g.\ to mine ranking data~\cite{HenzgenH14}, where association rules are used to find patterns within rankings.

We propose a method which combines the two approaches mentioned above~\cite{rebelosa2011,HenzgenH14}, because it can could contribute to a better understanding of the datasets mentioned above.
We define Pairwise Association Rules (PAR) as association rules with one or more pairwise comparisons in the consequent.
In this work we present an approach to identify PAR and analyze the findings in two real world datasets.%\csimportant{be consistent: use either PAR or PARs for the plural}

By decomposing rankings into the unitary preference relation i.e. \emph{pairwise comparisons}, we can look for sub-ranking patterns.
From which, as explained before, we expect to find more frequent patterns than with complete rankings.
%To keep it simple, we focus on a descriptive approach.
%Hence, we hope that PAR can help predictive methods improve their predictive accuracy by giving more insights of the data.

LRAR and PARs can be regarded as a specialization of general association rules that are obtained from data containing preferences, which we refer to as \emph{Preference Rules}.
These two approaches are complementary in the sense that they can give different insights from preference data.
We use LRAR and PAR in this work as predictive and descriptive models, respectively.

%4. summary of empirical tests to prove it

The paper is organized as follows:
Sections~\ref{sec:ar} and\ref{sec:labelranking} introduce the task of association rule mining and the label ranking problem, respectively;
Section~\ref{sec:arlr} describes the Label Ranking Association Rules and Section~\ref{sec:par} the Pairwise Association Rules proposed here; Section~\ref{sec:results} presents the experimental setup and discusses the results;
finally, Section~\ref{sec:conclusion} concludes this paper.

%%%%%%%%%%%%%%%%%%%%%%%%%%%%%%%%%%%%%%%%%%%%%%%%%%%%%%%%%%%%
\section{Association Rule Mining}
\label{sec:ar}

An association rule (AR) is an implication: $A \rightarrow C$ where $A \bigcap C = \emptyset$ and $A,C \subseteq \mathit{desc}\left(\mathbb{X}\right)$, where $\mathit{desc}\left(\mathbb{X}\right)$ is the set of descriptors of instances in the instance space $\mathbb{X}$, typically pairs $\left<attribute,value\right>$.
The training data is represented as $D=\{ \langle x_i \rangle\}$, $i=1,\ldots,\dataSize$, where $x_i$ is a vector containing the values $x^j_i , j=1,\ldots,\attSize$ of $\attSize$ independent variables, $\mathcal{A}$, describing instance $i$.
We also denote $ \mathit{desc}(x_i) $ as the set of descriptors of instance $ x_i $.
%\csunimportant{do you mean that $ \mathit{desc}(x_i) = \langle x_i \rangle$?}

\subsection{Interest measures}

There are many interest measures to evaluate association rules~\cite{Omiecinski03}, but typically they are characterized by \emph{support} and \emph{confidence}.
Here, we summarize some of the most common, assuming a rule $A \rightarrow C$ in $D$.

\paragraph{Support} percentage of the instances in $D$ that contain $A$ and $C$:
\[
\mathit{sup}\left(A \rightarrow C\right) = \frac{ \# \{x_i | A \cup C \subseteq \mathit{desc}(x_i), x_i \in D \} }{\dataSize}
\]

\paragraph{Confidence} percentage of instances that contain $C$ from the set of instances that contain $A$:
\[
\mathit{conf}\left(A \rightarrow C\right) = \frac{sup\left(A \rightarrow C\right)}{sup\left(A\right)}
\]

\paragraph{Coverage} proportion of examples in $D$ that contain the antecedent of a rule: \emph{coverage}~\cite{HalkidiV05}:
\[
\mathit{coverage}\left(A \rightarrow C\right) = sup\left(A\right)
\]
We say that a rule $A \rightarrow C$ covers an instance $x$, if $A \subseteq \mathit{desc}\left(x\right)$. 

\paragraph{Lift} measures the independence of the consequent, $C$, relative to the antecedent, $A$:
\[
	\mathit{lift}\left(A \rightarrow C\right)
	=
	\frac{sup(A \rightarrow C)}{sup(A) \cdot sup(C)}
\]
Lift values vary from 0 to $+\infty$.
If $A$ is independent from $C$ then $\mathit{lift}(A \rightarrow C)\sim1$.
%Considering that rules with high confidence can occur by chance~\cite{AzevedoJ07}

%However, $\mathit{lift}$ does not take into account the rule direction.

\subsection{APRIORI Algorithm}

The original method for induction of AR is the APRIORI algorithm, proposed in 1994 \cite{Agrawal1994}.
APRIORI identifies all AR that have support and confidence higher than a given minimal support threshold ($\mathit{minsup}$) and a minimal confidence threshold ($\mathit{minconf}$), respectively.
Thus, the model generated is a set of AR, $\mathcal{R}$, of the form $A \rightarrow C$, where $A,C \subseteq \mathit{desc}\left(\mathbb{X}\right)$, and $sup(A \rightarrow C) \geq \mathit{minsup}$ and $\mathit{conf}(A \rightarrow C) \geq \mathit{minconf}$.
For a more detailed description see \cite{Agrawal1994}.

Despite the usefulness and simplicity of APRIORI, it runs a time consuming candidate generation process and needs substantial time and memory space, proportional to the number of possible combinations of the descriptors.
Additionally it needs multiple scans of the data and typically generates a very large number of rules.
Because of this, many alternative methods were previously proposed, such as hashing~\cite{Park1995}, dynamic itemset counting~\cite{Brin1997}, parallel and distributed mining~\cite{Park1995b} and mining integrated into relational database systems~\cite{Sarawagi1998}.

In contrast to itemset-based algorithms, which compute frequent itemsets and rule generation in two steps, there are rule-based approaches such as FP-Growth (Frequent pattern growth method)~\cite{HanPYM04}.
This means that, rules are generated at the same time as frequent itemsets are computed.
%Rule based approaches allow for different pruning methods.

\subsection{Pruning}

 AR algorithms typically generate a large number of rules (possibly tens of thousands), some of which represent only small variations from others.
This is known as the rule explosion problem~\cite{bayardo2000constraint} which should be dealt with by pruning mechanisms.
Many rules must be discarded for computational and simplicity reasons.

Pruning methods are usually employed to reduce the amount of rules without reducing the quality of the model.
For example, an AR algorithm might find rules for which the confidence is only marginally improved by adding further conditions to their antecedent.%\cscritical{so?\ldots}
Another example is when the consequent $C$ of a rule $A \rightarrow C$ has the same distribution independently of the antecedent $A$.
In these cases, we should not consider these rules as meaningful.

\paragraph{Improvement}
A common pruning method is based on the improvement that a refined rule yields in comparison to the original one~\cite{bayardo2000constraint}.
The \emph{improvement} of a rule is defined as the smallest difference between the confidence of a rule and the confidence of all sub-rules sharing the same consequent:
\[
	\mathit{imp}(A \rightarrow C) = min(\forall A^\prime \subset A, \mathit{conf}(A \rightarrow C) - \mathit{conf}(A^\prime \rightarrow C))
\]

As an example, if one defines a minimum improvement $ \mathit{minImp}=1\% $, the rule $ A^\prime \rightarrow C $ will be kept if $ \mathit{conf}(A^\prime \rightarrow C) - \mathit{conf}(A \rightarrow C) \geq 1\% $, where $ A \subset A^\prime $.
%As an example, let us define a minimum improvement $ \mathit{minImp}=1\% $.
%A rule $ beer \wedge bread \rightarrow diapers $ with $91.5\%$ confidence will be discarded, if the simpler rule $beer \rightarrow diapers$ has confidence of $91\%$.

If $\mathit{imp}(A \rightarrow C) > 0$ we say that $A \rightarrow C$ is a productive rule.

\paragraph{Significant rules}

Another way to prune non productive rules is to use statistical tests \cite{Webb06}.
A rule is \emph{significant} if the confidence improvement over all its generalizations is statistically significant.
The rule $A \rightarrow C$ is significant if $\forall A^\prime \rightarrow C, A^\prime \subset A$ the difference $\textit{conf}\left(A \rightarrow C\right) - \textit{conf}\left(A^\prime \rightarrow C\right)$ is statistically significant for a given significance level ($\alpha$).

%%%%%%%%%%%%%%%%%%%%%%%%%%%%%%%%%%%%%%%%%%%%%%%%%%%%%%%%%%%%
\section{Label Ranking}
\label{sec:labelranking}

%A total order (or "totally ordered set," or "linearly ordered set") is a set plus a relation on the set (called a total order) that satisfies the conditions for a partial order plus an additional condition known as the comparability condition. A relation <= is a total order on a set S ("<= totally orders S") if the following properties hold.
%1. Reflexivity: a<=a for all a in S.
%2. Antisymmetry: a<=b and b<=a implies a=b.
%3. Transitivity: a<=b and b<=c implies a<=c.
%4. Comparability (trichotomy law): For any a,b in S, either a<=b or b<=a.
%The first three are the axioms of a partial order, while addition of the trichotomy law defines a total order.
% in Wolfram

In \ac{LR}, given an instance $x$ from the instance space $\mathbb{X}$, the goal is to predict the ranking of the labels $\mathcal{L} = \{\lambda_1,\ldots,\lambda_\rankingSize\}$ associated with $x$~\cite{Chenga}.
A ranking can be represented as a \emph{strict total order} over $\mathcal{L}$, defined on the permutation space $\Omega$.

The \ac{LR} task is similar to the classification task, where instead of a class we want to predict a ranking of labels.
As in classification, we do not assume the existence of a deterministic $\mathbb{X} \rightarrow\Omega$ mapping.
Instead, every instance is associated with a \emph{probability distribution} over $\Omega$~\cite{cheng2009icml}.
This means that, for each $x \in \mathbb{X}$, there exists a probability distribution $\mathcal{P}( \cdot | x)$ such that, for every $\pi \in \Omega$, $\mathcal{P}(\pi | x)$ is the probability that $\pi$ is the ranking associated with $x$.
The goal in \ac{LR} is to learn the mapping $\mathbb{X} \rightarrow \Omega$.
The training data contains a set of instances $D=\{ \langle x_i,\pi_i \rangle
\}$, $i=1,\ldots,\dataSize$, where $x_i$ is a vector containing the values $x^j_i , j=1,\ldots,\attSize$ of $\attSize$ independent variables, $\mathcal{A}$, describing instance $i$ and $\pi_i$ is the corresponding target ranking.

The rankings can be either total or partial orders.

\paragraph{Total orders}
A \emph{strict total order} over $\mathcal{L}$ is defined as:\footnote{For convenience, we say \emph{total order} but in fact we mean a \emph{totally ordered set}. Strictly speaking, a \emph{total order} is a binary relation.}
\[
	\{\forall \left(\lambda_a, \lambda_b\right) \in \mathcal{L} | \lambda_a \succ \lambda_b \vee \lambda_b \succ \lambda_a \}
\]
which represents a \emph{strict ranking}~\cite{VembuG10}, a \emph{complete ranking}~\cite{FurnkranzH10}, or simply a \emph{ranking}.
A strict total order can also be represented as a permutation $\pi$ of the set $\{1, \ldots, \rankingSize\}$, such that $\pi(a)$ is the position, or \emph{rank}, of $\lambda_a$ in $\pi$.
For example, the \emph{strict total order} $\lambda_1 \succ \lambda_2 \succ \lambda_3 \succ \lambda_4$ can be represented as $\pi=\left(1,2,3,4\right)$.

However, in real-world ranking data, we do not always have clear and unambiguous preferences, i.e.\ strict total orders \cite{BrandenburgGH13}.
Hence, sometimes we have to deal with \emph{indifference} and \emph{incomparability}.
For illustration purposes, let us consider the scenario of elections, where a set of $\dataSize$ voters vote on $\rankingSize$ candidates.
If a voter feels that two candidates have identical proposals, then these can be expressed as indifferent so they are assigned the same rank (i.e.\ a tie).

To represent ties, we need a more relaxed setting, called \emph{non-strict total orders}, or simply \emph{total orders}, over $\mathcal{L}$, by replacing the binary strict order relation, $\succ$, with the binary partial order relation, $\succeq$:
\[
	\{\forall \left(\lambda_a, \lambda_b\right) \in \mathcal{L} | \lambda_a \succeq \lambda_b \vee \lambda_b \succeq \lambda_a\}
\]
These non-strict total orders can represent \emph{partial rankings} (rankings with ties) \cite{VembuG10}.
For example, the \emph{non-strict total order} $\lambda_1 \succ \lambda_2 = \lambda_3 \succ \lambda_4$ can be represented as $\pi=\left(1,2,2,3\right)$.

Additionally, real-world data may lack preference data regarding two or more labels, which is known as \emph{incomparability}. 
Continuing with the elections example, the lack of information about one or two of the candidates, $\lambda_a$ and $\lambda_b$, leads to incomparability, $\lambda_a \perp \lambda_b$.
In other words, the voter cannot decide whether the candidates are equivalent or select one as the preferred, because he does not know the candidates.
Incomparability should not be confused with intrinsic properties of the objects, as if we are comparing apples and oranges.
Instead, it is like trying to compare two different types of apple without ever having tried either. In this cases, we can use \emph{partial orders}.

\paragraph{Partial orders}

Similarly to \emph{total orders}, there are \emph{strict} and \emph{non-strict partial orders}.
Let us consider the \emph{non-strict partial orders} (which can also be referred to as \emph{partial orders}) over $\mathcal{L}$:
%\[
%	\{\left( \lambda_a, \lambda_b\right) \in \mathcal{L} | \lambda_a \succeq \lambda_b \}
%\]
\[
	\{\forall \left(\lambda_a, \lambda_b\right) \in \mathcal{L} | \lambda_a \succeq \lambda_b \vee \lambda_b \succeq \lambda_a\ \vee \lambda_a \perp \lambda_b \}
\]
We can represent partial orders with \emph{subrankings}~\cite{HenzgenH14}.
For example, the \emph{partial order} $\lambda_1 \succ \lambda_2 \succ \lambda_4$ can be represented as $\pi=\left(1,2,0,4\right)$, where 0 represents $\lambda_1, \lambda_2, \lambda_4 \perp \lambda_3$.

%We use the same notation from order theory as in \cite{VembuG10}.
%A binary relation is defined as $\succ$ on a finite set
%A binary relation on a (finite) set ̇ is a partial order if is asymmetric
%(a b ) :b a) and transitive (a b ^ b c ) a c). The pair . ̇; / is
%then called a partially ordered set (or poset).
%We denote the set f.u; v/ 2 ̇ j u vg by p./ and the set of all partial orders
%over ̇ by P ̇ . Note that every partially ordered set . ̇; / defines a directed acyclic
%graph G D . ̇; p.//. This graph is also called as preference graph in the label
%ranking literature.
%A partially ordered set . ̇; / such that 8u; v 2 ̇ W u v _ v u is a totally
%ordered set and is called a total order, a linear order, a strict ranking (or simply
%ranking), or a permutation. A partial ranking is a total order with ties.
%A partial order 0 extends a partial order on the same ̇ if u v ) u 0 v.
%An extension 0 of a partial order is a linear extension if it is totally ordered
%i.e., a total order 0 is a linear extension of a partial order if 8u; v 2 ̇,
%u v ) u 0 v). A collection of linear orders i realizes a partial order if
%8u; v 2 ̇; u v , .8i W u i v/. We denote this set by `./. The dual of a
%partial order is the partial order N with 8u; v 2 ̇ W uN v , v u.
%\cite{VembuG10}

%\subsubsection{SubRankings}
%\cite{HwangL72}

%%%%%%%%%%%%%%%%%%%%%%%%%%%%%%%%%%%%%%%%%%%%%%%%%%%%%%%%%%%%
\subsection{Methods}

%\csunimportant{is it possible to make a statement of the sort: ``For more information on label ranking learning methods, more information ca be found in\ldots''}
Several learning algorithms were proposed for modeling label ranking data in recent years.
These can be grouped as decomposition-based or direct.
\emph{Decomposition-based methods} divide the problem into several simpler problems (e.g., multiple binary problems).
An example is ranking by pairwise comparisons~\cite{furnkranz+05} and mining rank data~\cite{HenzgenH14}.
\emph{Direct methods} treat the rankings as target objects without any decomposition.
Examples of that include decision trees~\cite{todorovski+02,cheng2009icml}, \emph{k}-Nearest Neighbors~\cite{brazdil+03,cheng2009icml} and the linear utility transformation~\cite{har-peled+02,dekel2003}.
This second group of algorithms can be divided into two approaches.
The first one contains methods that are based on statistical distributions of rankings (e.g.~\cite{cheng2009icml}), such as Mallows~\cite{lebanon+02b}, or Plackett-Luce~\cite{ChengDH10}.
The other group of methods are based on measures of similarity or correlation between rankings (e.g.~\cite{todorovski+02,aiguzhinov+10}).

LR-specific preprocessing methods have also been proposed, e.g. MDLP-R~\cite{rebelosa2013} and EDiRa~\cite{rebelosa2016}.
Both are \emph{direct methods} and based on measures of similarity.
Considering that supervised discretization approaches usually provide better results than unsupervised methods~\cite{Dougherty1995}, such methods can be of a great importance in the field.
In particular, for AR-like algorithms, such as the ones proposed in this work, which are typically not suitable for numerical data.

For more information on label ranking learning methods, more information ca be found in~\cite{plbook}.

%%%%%%%%%%%%%%%%%%%%%%%%%%%%%%%%%%%%%%%%%%%%%%%%%%%%%%%%%%%%
\subsubsection{Label Ranking by Learning Pairwise Preferences}
\label{sec:rpc}

Ranking by pairwise comparisons basically consists of reducing the problem of ranking into several classification problems.
In the learning phase, the original problem is formulated as a set of pairwise preferences problem. Each problem is concerned with one pair of labels of the ranking, $ \left( \lambda_i, \lambda_j \right) \in \mathcal{L},1\leq i < j \leq \rankingSize$. The target attribute is the relative order between them, $\lambda_i \succ \lambda_j$.
Then, a separate model $\mathcal{M}_{ij}$ is obtained for each pair of labels.
Considering $\mathcal{L} = \{\lambda_1,\ldots,\lambda_\rankingSize\}$, there will be $ \pairwiseSize = \frac{\rankingSize \left(\rankingSize-1\right)}{2} $ classification problems to model.

In the prediction phase, each model is applied to every pair of labels to obtain a prediction of their relative order. The predictions are then combined to derive rankings, which can be done in several ways.
The simplest is to order the labels, for each example, considering the predictions of the models $\mathcal{M}_{ij}$ as votes.
This topic has been well studied and documented~\cite{fodor1994fuzzy,Chenga}.

%
%\subsubsection{RPC}
%
%
%A learner $ \mathcal{M}_{ij} $ is supposed to learn a mapping of the form:
%\begin{equation}
%\label{eq:pw.pref1}
%x \rightarrow \left \{
%\begin{array}{rr}
%1 & \mbox{ if $ \lambda_i \succ \lambda_j $ } \\
%0 & \mbox{ if $ \lambda_j \succ \lambda_i $ }
%\end{array}
%\right .
%\end{equation}
%for all $ x\in \mathbb{X} $ which have information of preferences between $ \lambda_i $ and $ \lambda_j $.
%This mapping can be done by any common binary classifier.
%
%As a matter of choice, this can be easily adapted to deal with the interval $ \left[0, 1 \right] $.
%This will result in a \textit{valued preference relation} $ \mathcal{R_x} $ for every instance $ x \in \mathbb{X} $
%\begin{equation}
%\label{eq:pw.pref2}
%\mathcal{R_x}\left( \lambda_i, \lambda_j \right) \left \{
%\begin{array}{rr}
%\mathcal{M}_{ij} 	& \mbox{ if $ i < j $ } \\
%1-\mathcal{M}_{ij} 	& \mbox{ if $ i > j $ }
%\end{array}
%\right .
%\end{equation}

%Once obtained the predicted preference relations, $ \mathcal{R}_x $, we need to derive the rankings.

%Each label $ \lambda_i $ is ranked depending on the sum of the votes:
%\[
% S\left( \lambda_i \right)
% =
% \sum\limits_{\lambda_i \neq \lambda_j}^{} \mathcal{R}_x\left( \lambda_i, \lambda_j \right)
%\]

%\subsection{Preprocessing in Label Ranking}
%\todo{explain existing methods}

%%%%%%%%%%%%%%%%%%%%%%%%%%%%%%%%%%%%%%%%%%%%%%%%%%%%%%%%%%%%
\subsection{Evaluation}

Given an instance $x_i$ with label ranking $\pi_i$ and a ranking $\hat{\pi_i}$ predicted by a \ac{LR} model, several loss functions on $\Omega$ can be used to evaluate the accuracy of the prediction.
One such function is the number of discordant label pairs:
\[
	\mathcal{D}\left(\pi,\hat{\pi}\right)
	=
	\#\{(a,b) | \pi(a) > \pi(b) \wedge \hat{\pi}(a) < \hat{\pi}(b) \}
\]
If there are no discordant label pairs, the distance $\mathcal{D}=0$.
Alternatively, the function to define the number of concordant pairs is:
\[
	\mathcal{C}\left(\pi,\hat{\pi}\right)
	=
	\#\{(a,b) | \pi(a) > \pi(b) \wedge \hat{\pi}(a) > \hat{\pi}(b)\}
\]

\paragraph{Kendall Tau}
Kendall's $\tau$ coefficient~\cite{kendall1970rank} is the normalized difference between the number of concordant, $\mathcal{C}$, and discordant pairs, $\mathcal{D}$:
\[
\tau\left(\pi , \hat{\pi} \right) = \frac{\mathcal{C}-\mathcal{D}}{\frac{1}{2}\rankingSize\left(\rankingSize-1\right)}
\]
where $\frac{1}{2}\rankingSize\left(\rankingSize-1\right)$ is the number of possible pairwise combinations, ${\rankingSize \choose 2}$.
The values of this coefficient range from $[-1,1]$, where $\tau\left(\pi, \pi\right)=1$ if the rankings are equal and $\tau(\pi, \pi^{-1})=-1$ if $\pi^{-1}$ denotes the inverse order of $\pi$ (e.g.\ $\pi = (1,2,3,4)$ and $\pi^{-1} = (4,3,2,1)$).
Kendall's $\tau$ can also be computed in the presence of ties, using tau-b \cite{agresti2010analysis}.

An alternative measure is the Spearman's rank correlation coefficient~\cite{spearman04}.

\paragraph{Gamma coefficient}
If we want to measure the correlation between two partial orders (subrankings), or between total and partial orders, we can use the Gamma coefficient~\cite{kruskal1954}:
\[
\gamma\left(\pi , \hat{\pi} \right) = \frac{\mathcal{C}-\mathcal{D}}{\mathcal{C}+\mathcal{D}}
\]
Which is identical to Kendall's $\tau$ coefficient in the presence of strict total orders, because $\mathcal{C}+\mathcal{D} = \frac{1}{2}\rankingSize\left(\rankingSize-1\right)$.
%As an example, if we compute the gamma coefficient between $\pi_1=(1,2,3,4)$ and $\pi_2=(1,2,0,4)$.

\paragraph{Weighted rank correlation measures}
When it is important to give more relevance to higher ranks, a weighted rank correlation coefficient can be used.
They are typically adaptations of existing similarity measures, such as $\rho_w$~\cite{costa+04}, which is based on Spearman's coefficient.

These correlation measures are not only used for evaluation estimation, they can be used within learning~\cite{rebelosa2011} or preprocessing~\cite{rebelosa2016} models.
Since Kendall's $\tau$ has been used for evaluation in many recent \ac{LR} studies~\cite{cheng2009icml,rebelosa2013}, we use it here as well.

\hfill \break
The accuracy of a label ranker can be estimated by averaging the values of any of the measures explained here, over the rankings predicted for a set of test examples. Given a dataset, $D=\{ \langle x_i,\pi_i \rangle\}$, $i=1,\ldots,\dataSize$, the usual resampling strategies, such as holdout or cross-validation, can be used to estimate the accuracy of a LR algorithm.

\section{Label Ranking Association Rules}
\label{sec:arlr}

Association rules were originally proposed for descriptive purposes.
However, they have been adapted for predictive tasks such as classification (e.g., \cite{liu1998integrating}).
Given that label ranking is a predictive task, the adaptation of AR for label ranking comes in a natural way.
A \emph{Label Ranking Association Rule} (LRAR) \cite{rebelosa2011} is defined as:
\[
A \rightarrow \pi
\]
where $A \subseteq \mathit{desc}\left(\mathbb{X}\right)$ and $\pi \in \Omega$.
Let $\mathcal{R}_{\pi}$ be the set of \emph{label ranking association rules} generated from a given dataset.
When an instance $x$ is covered by the rule $A \rightarrow \pi$, the predicted ranking is $\pi$.
A rule $r_\pi: A \rightarrow \pi, r_\pi \in \mathcal{R}_{\pi}$, covers an instance $x$, if $A \subseteq \mathit{desc}(x)$.

We can use the CAR framework\cite{liu1998integrating} for LRAR.
However this approach has two important problems.
First, the number of classes can be extremely large, up to a maximum of $\rankingSize!$, where $\rankingSize$ is the size of the set of labels, $ \mathcal{L}$.
%For instance, if the number of labels is 5, the number of permutations is $5!=120$.
This means that the amount of data required to learn a reasonable mapping $\mathbb{X} \rightarrow\Omega$ is unreasonably large.

The second disadvantage is that this approach does not take into account the differences in nature between label rankings and classes.
In classification, two examples either have the same class or not.
In this regard, label ranking is more similar to regression than to classification.
In regression, a large number of observations with a given target value, say 5.3, increases the probability of observing similar values, say 5.4 or 5.2, but not so much for very different values, say -3.1 or 100.2. This property must be taken into account in the induction of prediction models.
A similar reasoning can be made in label ranking.
Let us consider the case of a data set in which ranking $\pi_a=\left(1,2,3,4\right)$ occurs in 1\% of the examples.
Treating rankings as classes would mean that $P(\pi_a)=0.01$.
Let us further consider that the rankings $\pi_b=\left(1,2,4,3\right), \pi_c=\left(1,3,2,4\right)$ and $\pi_d=\left(2,1,3,4\right)$, which are obtained from $\pi_a$ by swapping a single pair of adjacent labels, occur in 50\% of the examples.
Taking into account the stochastic nature of these rankings \cite{cheng2009icml}, $P(\pi_a)=0.01$ seems to underestimate the probability of observing $\pi_a$.
In other words it is expected that the observation of $\pi_b$, $\pi_c$ and $\pi_d$ increases the probability of observing $\pi_a$ and vice-versa, because they are similar to each other. 

This affects even rankings which are not observed in the available data.
For example, even though a ranking is not present in the dataset it would not be entirely unexpected to see it in future data.
This also means that it is possible to compute the probability of unseen rankings.

To take all this into account, similarity-based interestingness measures were proposed to deal with rankings \cite{rebelosa2011}.

\subsection{Interestingness measures in Label Ranking}
\label{sec:inter_mes}

As mentioned before, because the degree of similarity between rankings can vary, similarity-based measures can be used to evaluate LRAR.
These measures are able to distinguish rankings that are \emph{very similar} from rankings that are very \emph{very distinct}.
In practice, the measures described bellow can be applied to existing rule generation methods~\cite{rebelosa2011} (e.g.\ APRIORI~\cite{Agrawal1994}).

\paragraph{Support}

The support of a ranking $\pi$ should increase with the observation of similar rankings and that variation should be proportional to the similarity.
Given a measure of similarity between rankings $s(\pi_a, \pi_b)$, we can adapt the concept of support of the rule $A \rightarrow \pi$ as follows:
\[
sup_{lr}(A \rightarrow \pi)
=
\frac{\displaystyle\sum\limits_{i: A \subseteq \mathit{desc}(x_i)} s(\pi_i,\pi)}{n}
\]

Essentially, what we are doing is assigning a weight to each target ranking $\pi_i$ in the training data that represents its contribution to the probability that $\pi$ may be observed.
Some instances $ x_i \in \mathbb{X} $ give a strong contribution to the support count (i.e., 1), while others will give a weaker or even no contribution at all. 

Any function that measures the similarity between two rankings or permutations can be used, such as Kendall's $\tau$ \cite{kendall1970rank} or Spearman's $\rho$ \cite{spearman04}.
The function used here is of the form:
\begin{equation}
\label{eq:censored-sim}
s(\pi_a, \pi_b) = \left \{
\begin{array}{rr}
s'(\pi_a,\pi_b) & \mbox{ if $s'(\pi_a,\pi_b) \geq \theta$ } \\
0 & \mbox{ otherwise}
\end{array}
\right .
\end{equation}
where $s'$ is a similarity function.
This general form assumes that below a given threshold, $\theta$, is not useful to discriminate between different rankings, as they are so different from $\pi_a$. 
This means that, the support $sup_{lr}$ of $A \rightarrow \pi_a $ will be based only on the items of the form $\left<A,\pi_b \right>$, for all $\pi_b$ where $s'(\pi_a,\pi_b) > \theta)$.
%Again, 

Many functions can be used as $s'$. However, given that the loss function we aim to minimize is known beforehand, it makes sense to use it to measure the similarity between rankings.
Therefore, we use Kendall's $\tau$ as $s'$.

Concerning the threshold, given that anti-monotonicity can only be guaranteed with non-negative values~\cite{PeiHL01}, it implies that $\theta\geq0$.
Therefore we think that $\theta=0$ is a reasonable default value, because it separates between the positive and negative correlation between rankings.
 
Table~\ref{tbl:lr-data} shows an example of a label ranking dataset represented according to this approach.
Instance $\left(\{\mathcal{A}_1=L,\pi_3\}\right)$ (TID=1) contributes to the support count of ruleitem $\left< \{\mathcal{A}_1=L\}, \pi_3 \right> $ with 1, as expected.
However, that same instance, will also give a contribution of 0.33 to the support count of ruleitem $\left<\{\mathcal{A}_1=L\}, \pi_1 \right> $, given their ranking similarity.
On the other hand, no contribution to the support of ruleitem $\left<\{\mathcal{A}_1=L\}, \pi_2 \right> $ is given, because these rankings are clearly different.
This means that $sup_{lr}\left(\left< \{\mathcal{A}_1=L\}, \pi_3 \right>\right) =\frac{1 + 0.33}{3}$.

\begin{table}[tb]
\begin{center}
\caption{An example of a label ranking dataset.}
\label{tbl:lr-data}
\begin{tabular}{lcrrr}
\empty & \empty & $\pi_1$ & $\pi_2$ & $\pi_3$\\
TID & $\mathcal{A}_1$ & $(1,3,2)$ & $(2,1,3)$ & $(2,3,1)$\\
\hline
\textbf{1} & L & 0.33 & 0.00 & 1.00\\
\textbf{2} & L & 0.00 & 1.00 & 0.00\\
\textbf{3} & L & 1.00 & 0.00 & 0.33\\
\end{tabular}
\end{center}
\end{table}

\paragraph{Confidence}

The confidence of a rule $A \rightarrow \pi$ comes in a natural way if we replace the classical measure of support with the similarity-based $sup_{lr}$.
\[
\mathit{conf}_{lr}\left(A \rightarrow \pi\right)
=
\frac{\mathit{sup}_{lr}\left(A \rightarrow \pi\right)}{\mathit{sup}\left(A\right)}
\]

\paragraph{Improvement}

Improvement in association rule mining is defined as the smallest difference between the confidence of a rule and the confidence of all sub-rules sharing the same consequent \cite{bayardo2000constraint}.
In \ac{LR} it is not suitable to compare targets simply as equal or different (Section~\ref{sec:arlr}).
%The comparison of rules with different consequents can be done with the same approach explained earlier, using the test $ S^\prime\left( \pi,\pi^\prime \right) \geq \theta$.
Therefore, to implement pruning based on improvement for LR, some adaptation is required as well.
Given that the relation between target values is different from classification, as discussed in Section~\ref{sec:inter_mes}, we have to limit the comparison between rules with different consequents, if $ S'\left( \pi,\pi' \right) \geq \theta$.

Improvement for Label Ranking is defined as:
\[
	\mathit{imp}_{lr}(A \rightarrow \pi) = min(\mathit{conf}_{lr}(A \rightarrow \pi) - \mathit{conf}_{lr}(A^\prime \rightarrow \pi^\prime))
\]
for $\forall A^\prime \subset A$, and $\forall \left(\pi, \pi^\prime\right)$ where $ S^\prime\left(\pi^\prime,\pi\right) \geq \theta$.
As an illustrative example, consider the two rules $r_1: A_1 \rightarrow \left(1,2,3,4\right)$ and $r_2: A_2 \rightarrow \left(1,2,4,3\right)$, where $A_2$ is a superset of $A_1$, $ A_1 \subset A_2 $.
If $ S^\prime\left( \left( 1,2,3,4 \right) ,\left( 1,2,4,3 \right) \right) \geq \theta $ then $r_2$ will only be kept if, and only if, $ \mathit{conf}(r_1) - \mathit{conf}(r_2) \geq \mathit{minImp} $.

\paragraph{Lift}

The \emph{lift} measures the independence between the consequent and the antecedent of the rule~\cite{AzevedoJ07}.
%In particular for predictive models this measure has a big relevance because we are interested to know
The adaptation of lift for LRAR is straightforward since it only depends the concept of support, for which a version for LRAR already exists:
\[
	\mathit{lift}_{lr}(A \rightarrow \pi)
	=
	\frac{sup_{lr}(A \rightarrow \pi)}{sup(A) \cdot sup_{lr}(\pi)}
\]

%\subsubsection{Conviction}

%\subsubsection{Limitations on LRAR}
%
%\todo{still makes sense? mention that it cannot solve sushi?}
%Due to the similarity-based support used for \ac{LRAR} in some conditions, a significant increase of the support of the rules can be observed.
%This phenomenon depends on the similarity measure used and on the threshold $\theta$ but also depends on the dataset.
%For instance, if $\theta$ is very low and there are a lot of similar rankings in the data this can contribute for the increase of the support in many rules.
%
%Lets us suppose we use a ranking similarity measure $s$ that counts the proportion of concordant label pairs and $\theta=0$.
%If we are measuring the support of a rule:
%\[
%r_{\pi}^{a}: A \rightarrow \pi_a
%\]
%Any ranking $\pi$ associated with $A$ that has at least one concordant label pair with $\pi_a$ will contribute for the support of $r_{\pi}^a$.
%This can originate the generation of rules with high support.
%One way to deal with this problem is to adjust the $\theta$ parameter or to increase the $\mathit{minsup}$.

\subsection{Generation of LRAR}

Given the adaptations of the interestingness measures proposed, the task of learning LRAR can be defined essentially in the same way as the task of learning AR, i.e. to identify the set of LRAR that has a support and a confidence higher than the thresholds defined by the user.
More formally, given a training set $D=\{ \langle x_i,\pi_i \rangle \}, i=1,\ldots,\dataSize$, the algorithm aims to create a set of high accuracy rules $\mathcal{R}_{\pi}=\{ r_{\pi}: A \rightarrow \pi \}$ to cover a test set $T=\{ \langle x_j \rangle \}, j=1,\ldots,\testSize$.
If $\mathcal{R}_{\pi}$ does not cover some $x_j \in T$, a \emph{DefaultRanking} (Section~\ref{sssec:default_rules}) is assigned to it.

%Sticking to the original formulation, means that LRAR is a descriptive model.
%However, as mentioned before, in this work we use it in the context of prediction.

%\csimportant{I'm not sure that so much focus should be given to the implementation; Furthermore, does it make sense to have a single subsection?}

%\subsubsection{CAREN}
\subsubsection{Implementation of LRAR in CAREN}
%For the same reasons, the statistical based pruning procedures can also yield additional opportunities for pruning.

The association rule generator we are using is CAREN~\cite{Azevedo2010}.%
\footnote{\url{http://www4.di.uminho.pt/~pja/class/caren.html}}
CAREN implements an association rule algorithm to derive rule-based prediction models, like CAR and LRAR.
For Label Ranking datasets, CAREN derives association rules where the consequent is a complete ranking.

CAREN is specialized in generating association rules for predictive models and employs a bitwise depth-first frequent pattern mining algorithm.
Rule pruning is performed using a Fisher exact test~\cite{Azevedo2010}.
Like CMAR~\cite{Pei2010}, CAREN is a rule-based algorithm rather than itemset-based.
This means that, frequent itemsets are derived at the same time as rules are generated, whereas itemset-based algorithms carry out the two tasks in two separated steps.

Rule-based approaches allow for different pruning methods.
For example, let us consider the rule $A\rightarrow \lambda$, where $\lambda$ is the most frequent class in the examples covering $A$.
If $\mathit{sup}\left( A \rightarrow \lambda\right) < \mathit{minsup}$ then there is no need to search for a superset of $A$, $A^{*}$, since any rule of the form $A^{*} \rightarrow \lambda, A \subset A^{*} $ cannot have a support higher than $minsup$.

CAREN generates significant rules~\cite{Webb06}.
Statistical significance of a rule is evaluated using a Fisher Exact Test by comparing its support to the support of its direct generalizations.
The direct generalizations of a rule $A \rightarrow C$ are $\emptyset \rightarrow C$ and $\left(A\setminus\{a\}\right) \rightarrow C$ where $a$ is a single item.

The final set of rules obtained define the label ranking prediction model, which we can also refer as the \emph{label ranker}.
CAREN also employs prediction for strict rankings using \emph{consensus ranking} (Section~\ref{sec:prediction}), best rule, among others.

\subsection{Prediction}
\label{sec:prediction}

A very straightforward method to generate predictions using a label ranker is used.
The set of rules $\mathcal{R}_{\pi}$ can be represented as an ordered list of rules, by some user defined measure of relevance:
\[
 <r_{\pi_1},r_{\pi_2}, \ldots, r_{\pi_\rulesetSize}>
\]
As mentioned before, a rule $r^\ast_\pi: A^\ast \rightarrow \pi^\ast$ covers (or matches) an instance $x_i \in T$, if $A^\ast \subseteq \mathit{desc}(x_i)$.
If only one rule, $r^\ast_\pi$, matches $x_i$, the predicted ranking for $x_i$ is $\pi^\ast$.
However, in practice, it is quite common to have more than one rule covering the same instance $x_i$, $\mathcal{R}^\ast_{\pi}\left(x_j\right) \subseteq \mathcal{R}_{\pi}$.
In $\mathcal{R}^\ast_{\pi}\left(x_j\right)$ there can be rules with conflicting ranking recommendations.
There are several methods to address those conflicts, such as selecting the best rule, calculating the majority ranking, etc.
However, it has been shown that a ranking obtained by ordering the average ranks of the labels across all rankings minimizes the euclidean distance to all those rankings~\cite{kemeny+72}.
In other words, it maximizes the similarity according to Spearman's $\rho$ \cite{spearman04}.
This can be referred to as the \emph{average ranking}~\cite{brazdil+03}.

Given any set of rankings $\{\pi_i\}$ ($i=1,\ldots,\testSize$) with $\rankingSize$ labels, we compute the \emph{average ranking} as:
\begin{equation}
\label{eq:average_ranking}
 \overline{\pi}\left(j\right)
 =
 \frac{ \sum\limits_{i=1}^{\testSize} \pi_{i}\left(j\right) }{ \testSize }, j=1,\ldots,\rankingSize
\end{equation}
The average ranking $\overline{\pi}$ can be obtained if we rank the values of $\overline{\pi}\left(j\right), j=1,\ldots,\rankingSize$.
A weighted version of this method can be obtained by using the \emph{confidence} or \emph{support} of the rules in $\mathcal{R}^\ast_{\pi}\left(x_j\right)$ as weights.

\subsubsection{Default rules}
\label{sssec:default_rules}

As in classification, in some cases, the label ranker might not find any rule that covers a given instance $x_j$, so $\mathcal{R}^\ast_{\pi}\left(x_j\right) = \emptyset$.
To avoid this, we need to define a \emph{default rule}, $r_\emptyset$, which can be used in such cases:
\[
\{\emptyset\} \rightarrow \textit{default ranking}
\]

A \emph{default class} is also often used in classification tasks \cite{HanK2000}, which is usually the majority class of the training set $D$.
In a similar way, we could define the majority ranking as our \emph{default ranking}.
However, some label ranking datasets have as many rankings as instances, making the majority ranking not so representative.

As mentioned before, the \emph{average ranking} (Equation~\ref{eq:average_ranking}) of a set of rankings, minimizes the distance to all rankings in that set~\cite{kemeny+72}.
Hence we can use the \emph{average ranking} as the \emph{default ranking}.

%It has also been used as the baseline for LR problems (e.g.\ \cite{rebelo+08a}).

\subsection{Parameter tuning}
\label{sec:param_tun}

Due to the intrinsic nature of each different dataset, or even of the pre-processing methods used to prepare the data (e.g., the discretization method), the maximum $\mathit{minsup}/\mathit{minconf}$ needed to obtain a rule set $\mathcal{R}_{\pi}$, that covers all the examples, may vary significantly~\cite{LiuHM99a}.
The trivial solution would be, for example, to set $\mathit{minconf}=0$ which would generate many rules, hence increasing the coverage.
However, this rule would probably lead to a lot of uninteresting rules as well, as the model would overfit the data.
Then, our goal is to obtain a rule set $\mathcal{R}_{\pi}$ which gives maximal coverage while keeping high confidence rules.

Let us define $M$ as the coverage of the model i.e.\ the coverage of the set of rules $\mathcal{R}_{\pi}$.
Algorithm~\ref{al:partun} represents a simple, heuristic method to determine the $\mathit{minconf}$ that obtains the rule set such that a certain minimal coverage is guaranteed $minM$.

\begin{algorithm}
\caption{Confidence tuning algorithm}
\label{al:partun}
\begin{algorithmic}

\STATE Given $\mathit{minsup}$ and $\mathit{step}$
\STATE $\mathit{minconf}=100\%$
\WHILE{$M< minM$}
 \STATE $\mathit{minconf}=\mathit{minconf}-step$
 \STATE Run CAREN with ($\mathit{minsup}$,$\mathit{minconf}$) and determine $M$
\ENDWHILE

\RETURN $\mathit{minconf}$
\end{algorithmic}
\end{algorithm}

This procedure has the important advantage that it does not take into account the accuracy of the rule sets generated, thus reducing the risk of overfitting.

\section{Pairwise Association Rules}
\label{sec:par}

Association rules use a sets of descriptors to represent meaningful subsets of the data~\cite{Hastie2009}, hence providing an easy interpretation of the patterns mined.
Due to the intuitive representation, since its first application in the market basket analysis~\cite{Agrawal1993}, they have become very popular in data mining and machine learning tasks (Mining rankings \cite{HenzgenH14}, Classification \cite{liu1998integrating}, Label Ranking \cite{rebelosa2011}, etc).
%LRAR proved to be effective as a predictive model, however, as mentioned before, 
%Despite the existence of many competitive approaches in Label Ranking, Decision trees~\cite{todorovski+02,cheng2009icml}, \emph{k}-Nearest Neighbor~\cite{brazdil+03,cheng2009icml} or LRAR \cite{rebelosa2011}, problems with a large number of distinct rankings can be hard to predict.

LRAR proved to be an effective predictive model, however they are designed to find complete rankings.
Despite its similarity measures, which take into account possible ranking noise, it does not capture subranking patterns because it will always try to infer complete rankings.
On the other hand, association rules were used to find patterns within rankings~\cite{HenzgenH14}, however, they do not relate it with the independent variables.
Besides, in~\cite{HenzgenH14}, the consequent is limited to one pairwise comparison.

In this work, we propose a decomposition method to look for meaningful associations between independent variables and preferences (in the form of pairwise comparisons), the Pairwise Association Rules (PAR), which can be regarded as predictive or descriptive model.
We define PAR as:
\[
A \rightarrow \{\lambda_a \succeq \lambda_b \vee \lambda_a \perp \lambda_b\ \vee \lambda_a = \lambda_b | \lambda_a, \lambda_b \in \mathcal{L} \}
\]
where, as in the original AR paper \cite{Agrawal1994}, we allow rules with multiple items, not only in the antecedent but also in the consequent, i.e.\ PAR can have multiple sets of pairwise comparisons in the consequent.

Similarly to RPC (Section~\ref{sec:rpc}), we decompose the target rankings into pairwise comparisons.
Therefore, PAR can be obtained from data with strict rankings, partial rankings and subrankings.
\footnote{To derive the PAR, we added a pairwise decomposition method to the CAREN~\cite{Azevedo2010} software.}

Contrary to LRAR, we use the same interestingness measures that are also used in typical AR approaches, instead of the similarity-based versions defined for LR problems, i.e.\ \emph{sup}, \emph{conf}, etc.
This allows PAR to filter out non-frequent/interesting patterns and makes it more difficult to derive strict rankings.
When methods cannot find interesting rules with enough pairwise comparisons to define a strict ranking, partial rankings, subrankings or even with sets of disjoint pairwise comparisons can be found.
%In Association Rules mining, we filter out the rules below $\mathit{minsup}$, $\mathit{minconf}$, $\mathit{minlift}$, etc.
This is, interest measures are defining the borders between what the model will keep or abstain.

\emph{Abstention} is used in machine learning to describe the option to not make a prediction when the confidence in the output of a model is insufficient.
The simplest case is classification, where the model can abstain itself to make a decision~\cite{bartlettW08}.
In the label ranking task, a method that makes partial abstentions was proposed in \cite{cheng12abs}. 
A similar reasoning is used here both for predictive and descriptive models.

%Partial abstentions also make sense in PAR.
%Hence, the decision to abstain on certain preference orders is controlled with parameters, such as $\mathit{minconf}$ or $\mathit{lift}$.
%This way, we expect to obtain higher relevance PAR.

More formally, let us define $D=\{ \langle x_i, \pi_i \rangle\}, i=1,\ldots,\dataSize$ where $\pi_i$ can be a \emph{complete ranking}, \emph{partial ranking} or a \emph{sub-ranking}.
For each $\pi$ of size $\rankingSize$ we can extract up to $\pairwiseSize$ pairwise comparisons.
We consider 4 possible outcomes for each pairwise comparison:
\begin{itemize}
\item $\lambda_a \succeq \lambda_b$
\item $\lambda_b \succeq \lambda_a$
\item $\lambda_a = \lambda_b $ (indifference)
\item $\lambda_a \perp \lambda_b$ (incomparability)
\end{itemize}

%As an example, a PAR:
%\[
%	A \rightarrow \lambda_1 \succ \lambda_2 \wedge \lambda_2 \succ \lambda_3 \wedge \lambda_3 \succ \lambda_4
%\]
%can have $conf < minconf$.
%But, 

%The targets are represented as a set of binary variables which indicate whether the pairwise order relation is present ($\iota^{a,b} = TRUE$) or not ($\iota^{a,b} = FALSE$).
%In case of ties both $\iota^{a,b} = FALSE$ and $\iota^{b,a} = FALSE$.
%If incomparable $\iota^{a,b} = NA$.

As an example, a PAR can be of the form:
\[
	A \rightarrow \lambda_1 \succ \lambda_4 \wedge \lambda_3 \succ \lambda_1 \wedge \lambda_1 \perp \lambda_2
\]
The consequent can be simplified into $\lambda_3 \succ \lambda_1 \succ \lambda_4$ or represented as a subranking $\pi=\left(2,0,1,3\right)$.

%As mentioned before, there are $\frac{\rankingSize \left(k-1\right)}{2}$ possible combinations of pairwise comparisons, for rankings of size $\rankingSize$.
%In the case of the \emph{sushi} dataset, with 10 labels, we have $\frac{90}{2}=45$ pairwise comparisons.
%If we test all the combinations of preference rules with size 2 in the consequent, we have $\frac{45!}{2!\left( 45-2 \right)!}=990$.
%This means that for each antecedent we can test up to 990 combinations.
%
%To avoid this combinatorial problem we propose an heuristic approach.
%If we allow the creation of and at the same time improve interpretability of the consequent,
%Considering that PAR have orders in the consequent, we propose to couple the targets if and only if:
%\[
%x
%\]
%We only generate rules that can obtain incomplete orders.
%In this case we only have $\left(n-1\right)\left(n-2\right)$ tests, which in the case of sushi means 72.
%
%Taking into account that association rules with more than one consequent 

%\cscritical{shouldn't there be sections (or at least discussion) on implementation, prediction and default prediction, as for LRAR?}
%The contribution of PAR is very limited.

\section{Experimental Results}
\label{sec:results}
%-describe experimental setup (datasets, algorithms compared)

In this section we start by describing the datasets used in the experiments, then we introduce the experimental setup and finally present the results obtained.

\subsection{Datasets}

The data sets in this work were taken from KEBI Data Repository in the Philipps University of Marburg~\cite{cheng2009icml} (Table~\ref{tbl:datastats}).

To illustrate domain-specific interpretations of the results, we experiment with two additional datasets.
%\csunimportant{why do you refer to the new ones with capitalized first letter and not the KEBI ones?}
We use an adapted dataset from the 1999 COIL Competition~\cite{Bache+Lichman:2013}, Algae~\cite{rebelosa2016epm}, concerning the frequencies of algae populations in different environments.
The original dataset consisted of 340 examples, each representing measurements of a sample of water from different European rivers on different periods.
The measurements include concentrations of chemical substances like nitrogen (in the form of nitrates, nitrites and ammonia), oxygen and chlorine.
Also the pH, season, river size and its flow velocity were registered.
For each sample, the frequencies of 7 types of algae were also measured.
In this work, we considered the algae concentrations as preference relations by ordering them from larger to smaller concentrations.
Those with 0 frequency are placed in last position and equal frequencies are represented with ties.
Missing values in the independent variables were set to 0.

Finally, the Sushi preference dataset \cite{kamishima03}, which is composed of demographic data about 5000 people and sushi preferences is also used.
Each person sorted a set of 10 different sushi types by preference.
The 10 types of sushi, are a) shrimp, b) sea eel, c) tuna, d) squid, e) sea urchin, f) salmon roe, g) egg h) fatty tuna, i) tuna roll and j) cucumber roll.
Since the attribute names were not transformed in this dataset, we can make a richer analysis of it.

\begin{table}[tb]
\caption{Summary of the datasets}
\label{tbl:datastats}
\begin{center}
\small
 \begin{tabular}{lrrrrr}
\hline
Datasets& type & \#examples & \#labels & \#attributes & $U_{\pi}$ \\
\hline
%autorship 	& A & 841 & 4 & 70 & \\
bodyfat 	& B & 252 & 7 & 7 & 94\%\\
calhousing 	& B & 20,640 & 4 & 4 & 0.1\%\\
cpu-small 	& B & 8,192 & 5 & 6 & 1\%\\
elevators 	& B & 16,599 & 9 & 9 & 1\%\\
fried 		& B & 40,769 & 5 & 9 & 0.3\%\\
glass   	& A & 214 & 6 & 9 & 14\%\\
housing 	& B & 506 & 6  & 6 & 22\%\\
iris 		& A & 150 & 3 & 4 & 3\%\\
%pendigits 	& A & 10,992 & 10 & 16 & \\
segment 	& A & 2310 & 7 & 18 & 6\%\\
stock 		& B & 950 & 5 & 5 & 5\%\\
vehicle 	& A & 846 & 4 & 18 & 2\%\\
vowel 		& A & 528 & 11 & 10 & 56\%\\
wine  		& A & 178 & 3 & 13 & 3\%\\
wisconsin 	& B & 194 & 16 & 16 & 100\%\\
\hline
Algae (COIL) & \empty & 316 & 7 & 10 & 72\%\\
Sushi		& \empty & 5000 & 10 & 10 & 98\%\\
\hline
\end{tabular} 

\end{center}
\end{table}

Table~\ref{tbl:datastats} presents a simple measure of the diversity of the target rankings, the \emph{Unique Ranking's Proportion}, $U_{\pi}$.
$U_{\pi}$ is the proportion of distinct target rankings for a given dataset.
As a practical example, the \emph{iris} dataset has 5 distinct rankings for 150 instances, which results in $U_{\pi} = \frac{5}{150}\approx 3 \%$.
%This means that all the 150 rankings are duplicates of these 5.

\subsection{Experimental setup}

Continuous variables were discretized with two distinct methods:
(1) \emph{\ac{EDiRa}} (\cite{rebelosa2016}) and
(2) \emph{equal width} bins.
\emph{\ac{EDiRa}} is the state of the art supervised discretization method in Label Ranking, while \emph{equal width} is a simple, general method that serves as baseline.

The evaluation measure used in all experiments is Kendall's $\tau$.
A ten-fold cross-validation was used to estimate the value for each experiment.
The generation of \ac{LRAR} and PAR was performed with CAREN~\cite{Azevedo2010} which uses a depth-first based approach.

The confidence tuning Algorithm~\ref{al:partun} was used to set parameters.
We consider that $5\%$ seems a reasonable step value because the $\mathit{minconf}$ can be found in, at most, 20 iterations.
Given that a common value for the $\mathit{minsup}$ in \ac{AR} mining is $1\%$, we use it as default for all datasets.
We define the \emph{minM} as $95\%$ to get a reasonable coverage, and $\textit{minImp}=1\%$ to avoid rule explosion.

In terms of similarity functions, we use a normalized Kendall $\tau$ between the interval $[0,1]$ as our similarity function $s$ (Equation~\ref{eq:censored-sim}).

\subsection{Results with LRAR}

In the experiments described in this section we analyze the performance from different perspectives, \emph{accuracy}, \emph{number of rules} and \emph{average confidence} as the similarity threshold $\theta$ varies.
We expect to understand the impact of using similarity measures in the generation of LRAR and provide some insights about its usage.

LRAR, despite being based on similarity measures, are consistent with the classical concepts underlying association rules.
A special case is when $\theta=1$, where, as in CAR, only equal rankings are considered.
Therefore, by varying the threshold $\theta$  we also understand how similarity-based interest measures ($0\leq\theta<1$) contribute to the accuracy of the model, in comparison to frequency-based approaches ($\theta=1$).
%When $\theta=1$,  the distance function is 0-1 loss, hence the generation of LRAR equivalent to the generation of CAR.

We would also like to understand how some properties of the data relate the sensitivity to $\theta$.
We can extract two simple measures of ranking diversity from the datasets, the \emph{Unique Ranking's Proportion} ($U_{\pi}$), mentioned before, and the \emph{ranking entropy}~\cite{rebelosa2016}.

\subsubsection{Sensitivity analysis}

Here we analyze how the similarity threshold $\theta$ affects the accuracy, number and quality (in terms of confidence) of LRAR.

\paragraph{Accuracy}

\begin{figure}[thp]
     \centering
     \centerline{\includegraphics[scale=0.6]{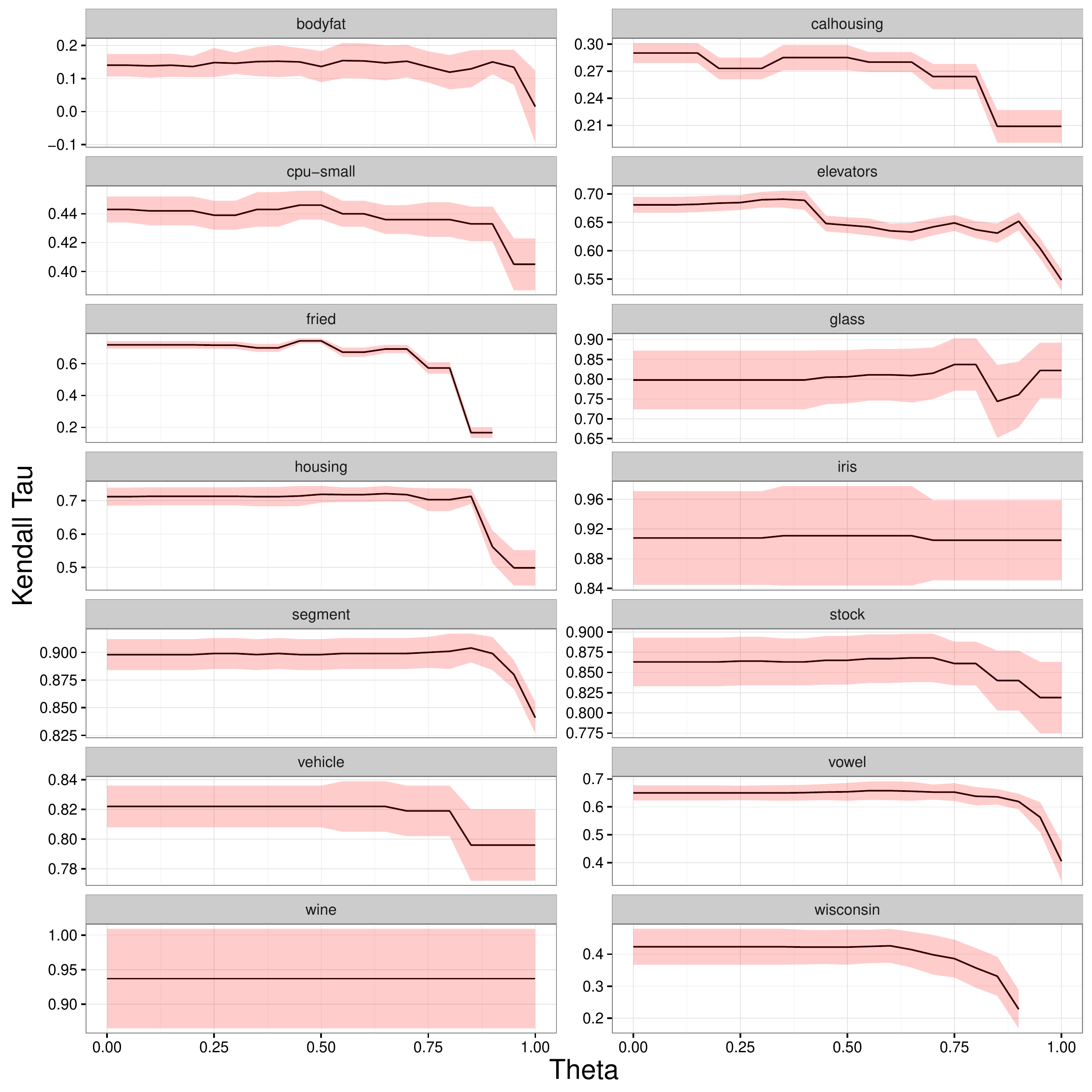}}
     \caption{Average accuracy (Kendall $\tau$) of CAREN as the $\theta$ varies}
     \label{graph:theta_tau}
\end{figure}

In Figure~\ref{graph:theta_tau} we can see the behavior of the accuracy of CAREN in terms of $\theta$.
It shows that, in general, there is a tendency for the accuracy to decrease as $\theta$ gets closer to 1.
This happens in 12 out of the 14 datasets analyzed.
On the other hand, in 9 out of 14 datasets, the accuracy is rather stable in the range $\theta \in [0,0.6]$.

%\cscritical{I think you need to say something more about the behavior you mention. You just discuss the range from 0 to 0.6 and the value of 1; what about the other values? and what is the take-away message for the users of the algorithm?}
If we take into consideration that the model ignores all similarities between rankings for $\theta=1$, the observed behavior seems to favor the similarity-based approach.
In line with that, two extreme cases can be seen with \emph{fried} and \emph{wisconsin} datasets, where CAREN was not able to find any \ac{LRAR} for $\theta = 1$.
\footnote{The \emph{default rule} was not used in these experiments because it is not related with $\theta$.}

\begin{figure}[thp]
     \centering
     \centerline{\includegraphics[scale=0.29]{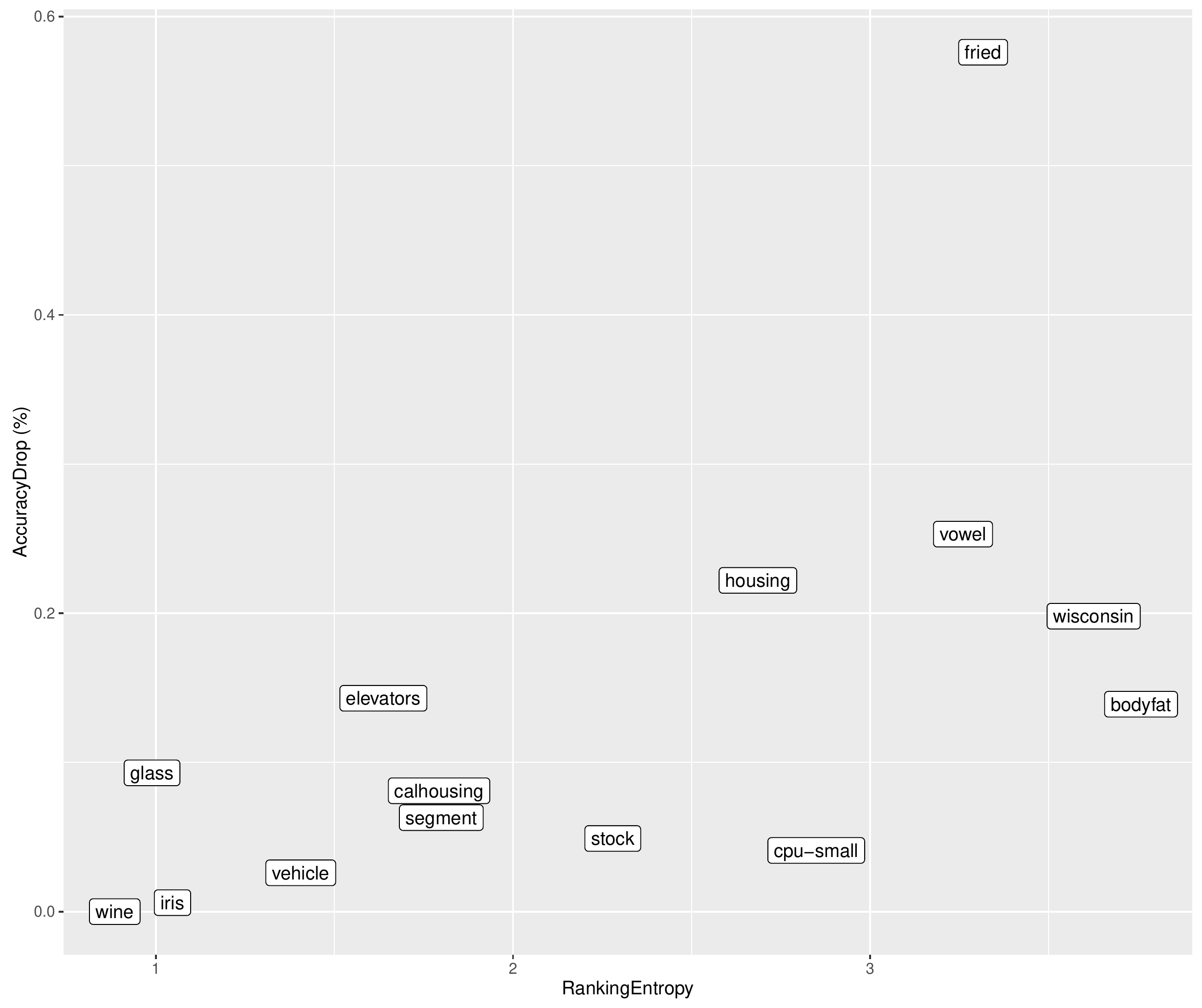}}
     \caption{Measured \emph{accuracy range} (Kendall $\tau$) of CAREN in comparison to ranking entropy.}
     \label{graph:urank_acc}
\end{figure}

Let us consider the \emph{accuracy range}, the maximum accuracy minus the minimum accuracy.
To find out which datasets are more likely to be affected by the choice of $\theta$, we can compare their ranking entropy with the measured \emph{accuracy range} from Figure~\ref{graph:theta_tau}.
%The three biggest differences are in \emph{housing}, \emph{vowel} and \emph{wisconsin} with $0.265$, $0.238$ and $0.197$ respectively.
%With an $U_{\pi}$ of $22\%$, $56\%$ and $100\%$, \emph{housing}, \emph{vowel} and \emph{wisconsin} represent 3 of the 4 highest $U_{\pi}$ (Table~\ref{tbl:datastats}).
In Figure~\ref{graph:urank_acc} we compare the accuracy range with the \emph{ranking entropy}~\cite{rebelosa2016}.
We can see that, the higher the entropy, the more the accuracy can be affected by the choice of $\theta$.

Results seem to indicate that, when mining LRAR in datasets with low ranking entropy, the choice of $\theta$ is not so relevant.
On the other hand, as the entropy gets bigger, a reasonable value should be $0 \leq \theta \leq 0.6$.

One interesting behavior can be found in the dataset \emph{fried}.
Despite the fact that it has a very low proportion of unique rankings, $U_{\pi}\left(\textit{fried}\right)=0.3\%$ (Table~\ref{tbl:datastats}) its entropy is quite high (Figure~\ref{graph:urank_acc}).
For this reason, it makes it more sensitive to $\theta$, as seen in Figure~\ref{graph:theta_tau}.
On the other hand, \emph{iris} and \emph{wine}, with very low entropy, seem unaffected by $\theta$.
%This means that, label ranking datasets, both with low $U_\pi$ and low ranking entropy, should have similar results if adapted to classification.

%\cscritical{figure is not easy to read}

\paragraph{Number of rules}

Ideally, we would like to obtain a small number of rules with high accuracy.
However, such a balance is not expected to happen frequently.
Ultimately, as accuracy is the most important evaluation criterion, if a reduction in the number of rules comes with a high cost in accuracy, it is better to have more rules.
Thus, it is important to understand how the number of LRAR varies with the similarity threshold $\theta$, while taking the impact in the accuracy of the model into account as well.

\begin{figure}[htp]
     \centering
     \centerline{\includegraphics[scale=0.6]{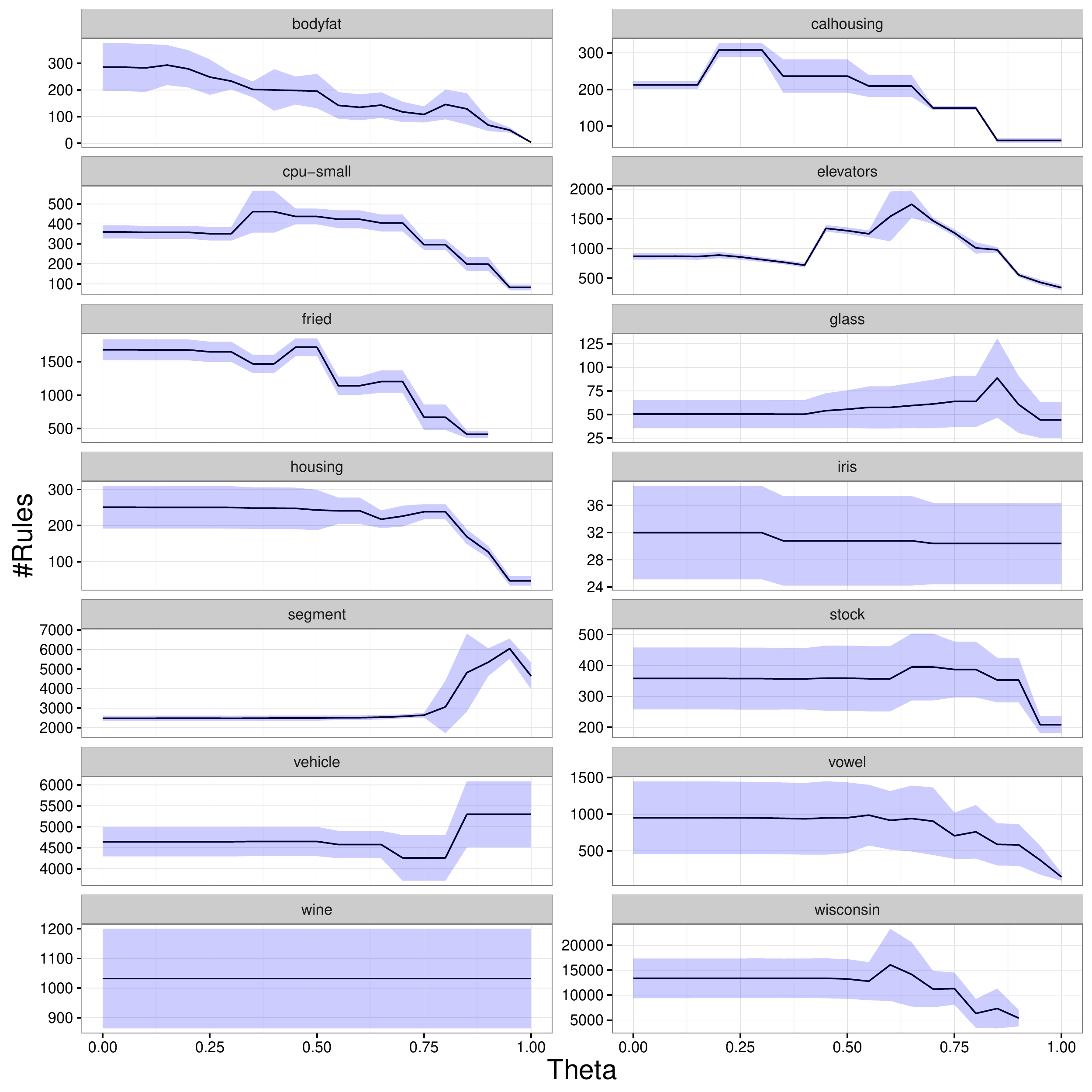}}
     \caption{Number of Label Ranking Association Rules generated by CAREN as the $\theta$ varies}
     \label{graph:theta_rules}
\end{figure}

In Figure~\ref{graph:theta_rules} we see how many LRAR are generated per dataset as $\theta$ varies.
The majority of the plots, 10 out of 14, show a decrease in the number of rules as $\theta$ gets closer to 1.
As discussed before, the accuracy in general also decreases as $\theta \geq 0.6$, so let us focus on $\theta \in [0,0.6]$.

In the interval $\theta \in [0,0.6]$, the number of rules generated is quite stable in 9 out of 14 datasets.
In the first half of this interval, $\theta \in [0,0.3]$, it is even more remarkable for 13 datasets.
%\csimportant{this is not clear from the plot: are we talking about bodyfat or calhousing? this indicates that you should quantify this variation, possibly in a simple way, by discussing the range}
%However, instead of a constant behavior, intuitively, we expect more rules when $\theta=0$ and much less when $\theta=1$.

We expect the number of rules to decrease as $\theta$ increases, however, results show that the number of rules does not decrease so much, especially for values up to 0.3.
This is due to the fact that $\theta$ is also used in the pruning step (Section~\ref{sec:inter_mes}), reducing the number of rules against which the improvement of an extension is measured and, thus, increasing the probability of an extension not being kept in the model.
This means that, $\textit{minImp}_{lr}$ is being effective in the reduction of LRAR.

As mentioned before, $\textit{imp}_{lr}\left(A \rightarrow \pi\right)$ not only compares rules $A^\prime \rightarrow \pi$ where $A^\prime \subset A$, but also rules $A \rightarrow \pi^\prime$ where $S^\prime\left(\pi^\prime,\pi\right) \geq \theta $.
In other words, with the $minImp_{lr}$ we are pruning LRAR with similar rankings too.
%One side effect is that, as we increase $\theta$, $minImp_{lr}$ loses a bit of pruning ``power''.
%\cscritical{It is not clear which observations is supporting this claim; I think that if you analyze the reduction between 0.3 and 0.6 you can support this claim; and if this is true, then in the next paragraph, you can claim that this result, combined with the previous analysis on accuracy, indicates that $\theta$ should be 0.6}

These results do not lead to any strong conclusions about the ideal value for $\theta$ regarding the number of rules.
However, they are in line with the previous analysis of \emph{accuracy}.

\paragraph{Minimum Confidence}

\begin{figure}[htp]
     \centering
     \centerline{\includegraphics[scale=0.6]{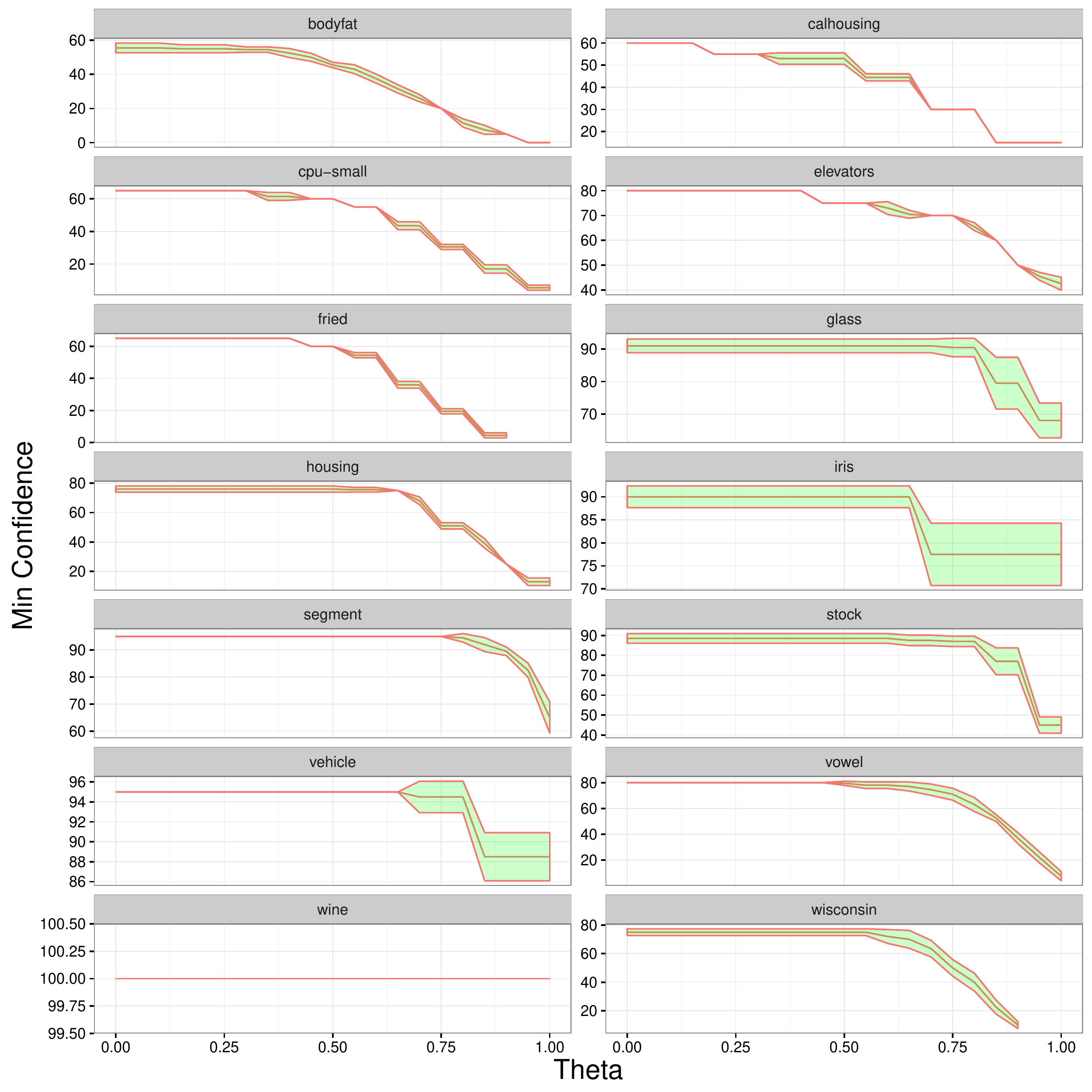}}
     \caption{Mininum confidence adjusted to CAREN as the $\theta$ varies}
     \label{graph:theta_minconf}
\end{figure}

As mentioned before, we use a greedy algorithm to automatically adjust the minimum confidence in order to reduce the number of examples that are not covered by any rule.
%\cscritical{if I understand correctly, ``per dataset'' is by definition; ``per $\theta$'' is a consequence of the empirical results; if this is true, this should be analyzed separately}
This means that the method has to adapt the value of $\mathit{minconf}$ per dataset per $\theta$, as seen in Figure~\ref{graph:theta_minconf}.

In general, the $\mathit{minconf}$ decreases in a monotonic way as $\theta$ increases.
As $\theta\approx 1$ the $\mathit{minconf}$ gets to its minimum with 13 out of 14 datasets, which is consistent with the accuracy plots (Figure~\ref{graph:theta_tau}).
%\csimportant{this contradicts the previous observation that I proposed}
This means that, if we want to generate rules with as much confidence as possible, we should use the minimum $\theta$, i.e.\ $\theta =0$.

%Since the confidence of rules indicate how certain the method is of its predictions, we can see it as an expectation measure of the model.
%In other words, the method is not so certain about the predictions as $\theta \approx 1$.

%We can make a comparison here between this phenomenon and over-fitting.
%To exemplify, lets consider a LRAR $A\rightarrow \pi$. (see definition of set of rules)
%If $A\rightarrow \pi$ can be more robust 

\paragraph{Support versus accuracy}

We vary the minimum support threshold, $\mathit{minsup}$, to test how it affects the accuracy of our learner.
A similar study has been carried out on CBA~\cite{iqbal2013comparison}.
Specifically, we vary the $\mathit{minsup}$ from $0.1\%$ to $10\%$, using a step size of $0.1\%$.
Due to the complexity of these experiments, we only considered the six smallest datasets.
\begin{figure}[htp]
     \centering
     \centerline{\includegraphics[scale=0.29]{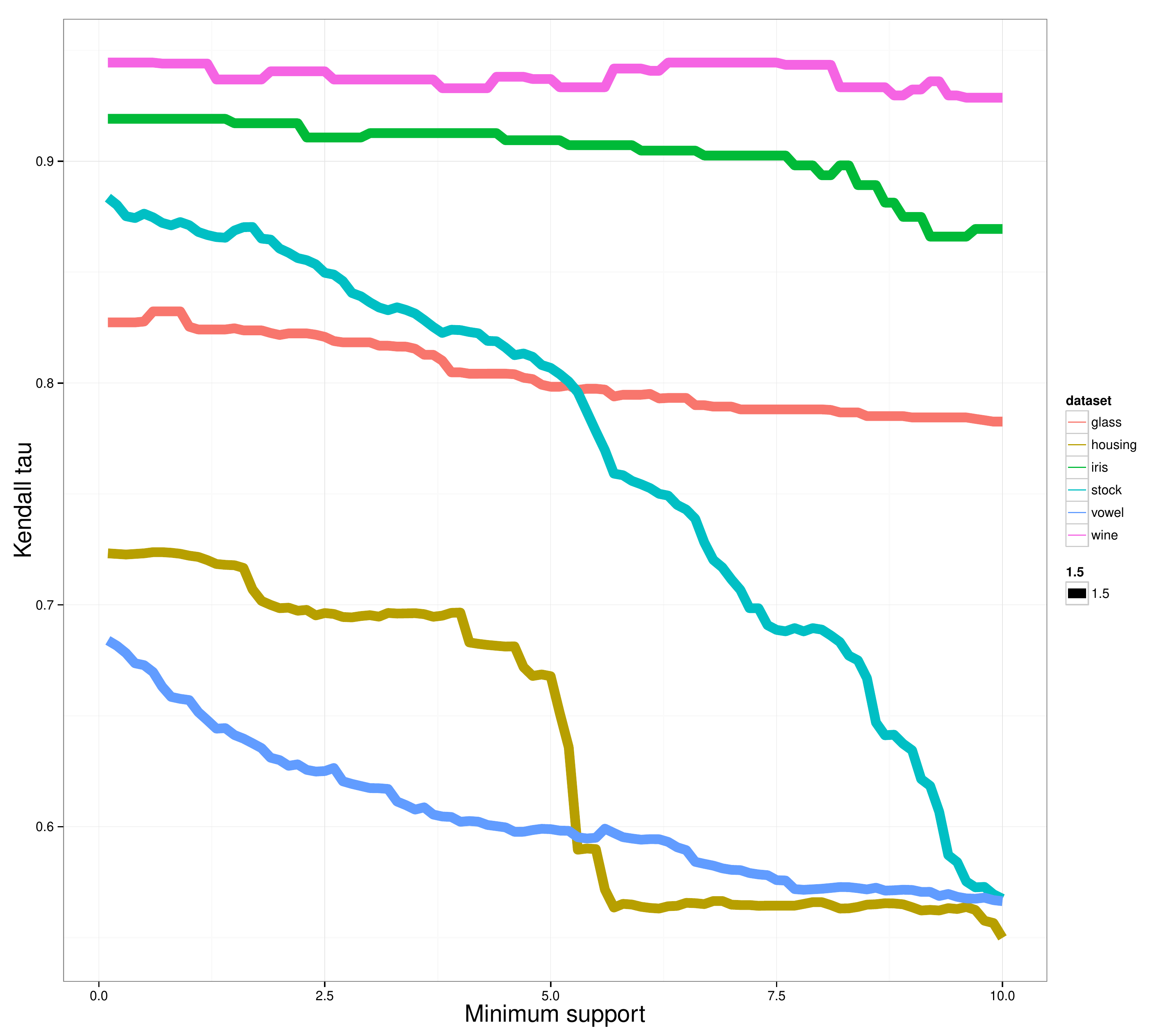}}
     \caption{Average accuracy (Kendall $\tau$) of CAREN as the $\mathit{minsup}$ varies.}
     \label{graph:sup_tau}
\end{figure}

In general, as we increase $\mathit{minsup}$ the accuracy decreases, which is a strong indicator that the support should be small (Figure~\ref{graph:sup_tau}).
All lines are monotonically decreasing, i.e.\ either the values remain constant or they decrease as $\mathit{minsup}$ increases.
%Moreover, all datasets analyzed have worst results when $\mathit{minsup}=10\%$ and the best when $\mathit{minsup}=0.1\%$.

From a different perspective, the changes are generally very small for $\mathit{minsup} \in [0.1\%,1.0\%]$.
Considering that lower $\mathit{minsup}$ generate potentially more rules, we recommend $\mathit{minsup} = 1\%$ as a reasonable value to start experiments with.

\paragraph{Discretization techniques}

To test the influence of the discretization method used, we performed the same analysis using a non-supervised discretization method, \emph{equal width}.

\begin{figure}[htp]
    \vspace*{-0.4in}
     \centering
     \centerline{\includegraphics[scale=0.25]{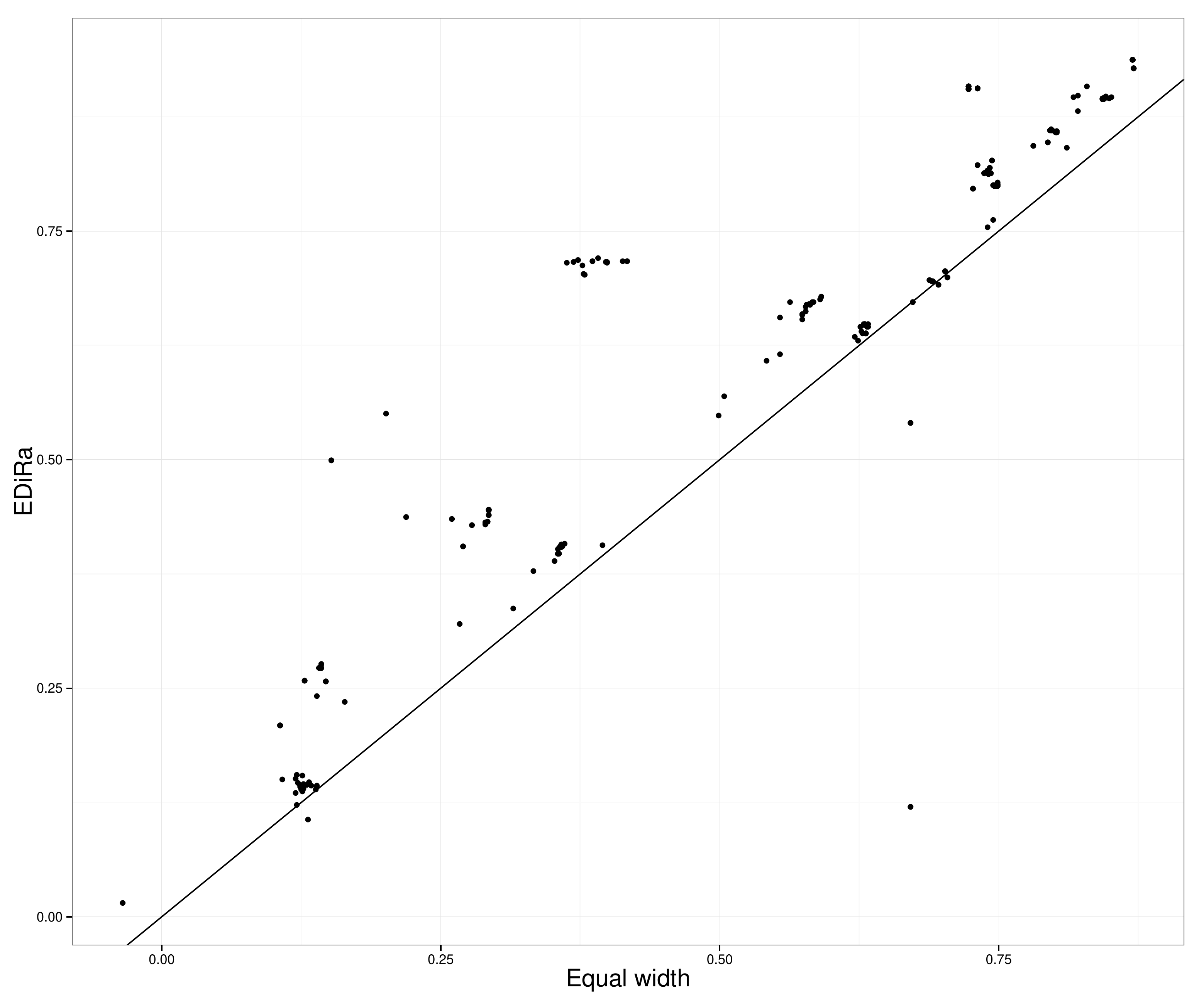}}
     \caption{Ranking accuracy (Kendall $\tau$) of CAREN after the discretization of data using \emph{equal width} and \emph{EDiRa}. This plot aggregates all the experiments carried out, concerning different issues, which means that each dataset is represented multiple times, with different parameter settings.}
     \label{graph:ew_edira}
\end{figure}

In general, the accuracy had the same behavior, as a function of $\theta$, as with \emph{EDiRa}, i.e.\ the results are highly correlated (Figure~\ref{graph:ew_edira}).
However, the supervised approach is consistently better.
%This overrules the influence of \emph{EDiRa} in the conclusions.

%Even though the accuracy is very similar using data discretized by both methods, it is slightly better when using the supervised approach (Figure~\ref{graph:ew_edira}).
These results add further evidence that \emph{EDiRa} is a suitable discretization method for label ranking~\cite{rebelosa2016}. 
%\cscritical{add $\tau$ to the axis labels}

Similar behavior was observed concerning the number of rules generated and the minimum confidence.

\paragraph{Summary}

It is well known that general, simple rules to set parameters of machine learning algorithms do not exist.
Nevertheless it is good to know where reasonable values lie.
Hence, we think that $\theta \in \left[0.5,0.6\right]$ and $\mathit{minsup}=1\%$ are good default values for LRAR with CAREN.
In terms of the discretization methods, our results confirm that a supervised approach, such as \emph{EDiRa}, is a good choice.

\subsection{Results with PAR}

In this work we use PAR, as a descriptive model, to find patterns concerning subsets of labels.
We focus in the descriptive task for two reasons.
One is to make the approach more simple and the other one is because this complements with the predictive LRAR approach.

The minimum support and confidence presented here are defining the abstention level of the model.
The $\mathit{minsup}$ and $\mathit{minconf}$ were adjusted manually to generate a small set of rules between 150 to 200.

In the generation of PAR, we set the minimum lift to $1.5$.
Despite that many interesting rules were found, due to space limitations we only present the most relevant.

\paragraph{Algae data}

Using the Algae dataset, we found 179 PARs with $\mathit{minsup}=2$ and $\mathit{minconf}=90$.
With $\mathit{sup}=2.2\%$ and $\mathit{conf}=100\%$ the rule with the highest lift (approx. 6) was:
\begin{align*}
	&\mathtt{River size}=\mathit{small} \wedge  
	\mathtt{pH} \ge 37.9 \wedge 
	\mathtt{Flow velocity} = \mathit{high} \wedge \\
	&\mathtt{Chloride} \ge 3.4 \wedge 
	\mathtt{Nitrates\&Ammonia} \ge 18.5 \\
	&\rightarrow 	L6\succ L2 \wedge L5\succ L7 \wedge L2\succ L7 
\end{align*}
The consequent of this rule can be represented as $L6\succ L2\succ L7 \wedge L5\succ L7$. 
Considering that the labels represent algae populations, this rule states that it is always true that, under these conditions, type 6 is more prevalent than type 2. It also states that type 7 is less prevalent than types 2, 5 and 6.

The second rule with highest lift, with $\mathit{sup}=3.1\%$ and $\mathit{conf}=91\%$ is:
\begin{align*}
	&\mathtt{Flow velocity}=\mathit{medium} \wedge \mathtt{Nitrates\&Ammonia} < 18.5 \wedge \\
	&\mathtt{Nitrogen as nitrates} < 7.9 \\
	&\rightarrow 	L1\succ L7 \wedge L7\succ L3
\end{align*}
The target of this rule is the partial ranking $L1\succ L7\succ L3$.

%The limits for support and confidence are interfering with the abstention level of the model.
If this PAR was used for prediction, the subranking $\pi=\left(1,0,3,0,0,0,2\right)$ would have been the prediction.

\paragraph{Sushi data}

When analyzing the sushi dataset we got 166 rules with $\mathit{minconf}=70\%$ and the $\mathit{minsup}=1\%$.
With a lift of $1.95$ the following rule was found:
\begin{align*}
	&\mathtt{Age interval}=15-19 \wedge \mathtt{Sex}=\mathit{Male} \wedge \mathtt{Lived in} = \mathit{Eastern\ Japan} \\
	&\rightarrow 	\mathtt{egg} \succ \mathtt{sea urchin} \wedge \mathtt{shrimp} \succ \mathtt{sea urchin}
\end{align*}

In the whole dataset, $37\%$ of the people show this relative preferences $\mathtt{egg} \succ \mathtt{sea urchin} \wedge \mathtt{shrimp} \succ \mathtt{sea urchin}$.
This PAR shows that this number almost double ($72\%$), if we consider males from Eastern Japan, aged between $15-19$.
%, prefer $\mathtt{egg}$ or $\mathtt{shrimp}$ to $\mathtt{sea urchin}$.

A related rule was also found concerning a different group of people, with different age and from a different region ($\mathit{sup}=1.1\%$, $\mathit{conf}=71.6\%$ and $\mathit{lift}=1.65$):
\begin{align*}
	&\mathtt{Age interval}=30-39 \wedge \mathtt{Sex}=\mathit{Male} \wedge \\
	&\mathtt{Lives in} = \mathit{Western\ Japan} \wedge \mathtt{Changed city} = \mathit{Yes} \\
	&\rightarrow 	\mathtt{sea urchin} \succ \mathtt{egg} \wedge \\
	&\mathtt{fatty tuna} \succ \mathtt{tuna roll} \wedge \\
	&\mathtt{tuna roll} \succ \mathtt{cucumber roll} \wedge \\
	&\mathtt{fatty tuna} \succ \mathtt{egg}
\end{align*}
This rule includes one relative preference found in this group, $\mathtt{sea urchin} \succ \mathtt{egg}$, which is the opposite to what was observed in the previous rule. 
Based on this information, we analyzed the data and found out that $75\%$ of people that live in Eastern Japan prefer $\mathtt{egg}$ to $\mathtt{sea urchin}$ while $84\%$ of people from Western Japan prefer $\mathtt{sea urchin}$ to $\mathtt{egg}$.

\section{Conclusions}
\label{sec:conclusion}

In this paper we address the problem of finding association patterns in label rankings.
We present an extensive empirical analysis on the behavior of a label ranking method, the CAREN implementation of Label Ranking Association Rules.
The performance was analyzed from different perspectives, \emph{accuracy}, \emph{number of rules} and \emph{average confidence}.
The results show that, similarity-based interest measures contribute positively to the accuracy of the model, in comparison to frequency-based approaches, i.e.\ when $\theta=1$.
%LRAR, despite being based on similarity measures, are consistent with the classical concepts underlying association rules.
%In fact, CAR can be regarded as a special case of LRAR, where the distance function is 0-1 loss, i.e.\ when $\theta=1$.
%
%Our experiments confirm the advantage of this approach by showing that, in most cases, the accuracy decreases as $\theta$ gets closer to 1.
%On the other hand, the best performance lies when $\theta$ is closer to zero.
%The results indicate that the contributions are more evident in datasets where the ranking entropy is high.
The results confirm that LRAR are a viable label ranking tool which helps solving complex label ranking problems (i.e. problems with high ranking entropy).
The results also enabled the identification of some values for the parameters of the algorithm that are good candidates to be used as default values.

Results also seem to indicate that, the higher the entropy, the more the accuracy can be affected by the choice of $\theta$.
An user can measure the ranking entropy of a dataset beforehand and adjust $\theta$ accordingly.

%The accuracy of LRAR was also studied from a \emph{minsup} perspective.
%With no surprise, the best results were obtained for smaller \emph{minsup}, specially when $\mathit{minsup} \in [0.1\%,1.0\%]$. 

% All the numeric variables we discretized using a supervised discretization method for label ranking, \emph{EDiRa}.
% The same experiments were also performed with discretization from an unsupervised method, \emph{equal width}.
% Given that the behavior of CAREN did not change dramatically from one to the other, the results from the latter were not presented here in detail.
% In general, the supervised discretization makes CAREN more accurate.

Additionally, we propose Preference Association Rules (PAR), which are association rules where the consequent represents multiple pairwise preferences.
We illustrated the usefuleness of this approach to identify interesting patterns in label ranking datasets, which cannot be obtained with LRAR.

% PAR on the other hand, successfully found interesting sub-ranking patterns in two dataset, \emph{Algae} and \emph{Sushi}.
% Due to space limitations, only a small portion of rules was presented and discussed in detail to show the potential of our approach.
% As a proof of concept this seems a very promising approach.
In future work, we will use PAR for predictive tasks. 
%This requires the use of methods to aggregate the pairwise comparisons in the consequent.

%-some ideas for future work
% This work contributed with a comprehensive analysis of the performance of LRAR.
% In the end, we hope that this could serve as a guideline on how to pick a suitable $\theta$ for future label ranking datasets.

\section*{Acknowledgments}

This work is financed by the ERDF - European Regional Development Fund through the Operational Programme for Competitiveness and Internationalization - COMPETE 2020 Programme within project POCI-01-0145-FEDER-006961, and by National Funds through the FCT - Funda\c{c}\~{a}o para a Ci\^{e}ncia e a Tecnologia (Portuguese Foundation for Science and Technology) as part of project  UID/EEA/50014/2013.

\acrodef{LR}[LR]{Label Ranking}
\acrodef{AR}{Association Rule}
\acrodef{LRAR}[LRAR]{Label Ranking Association Rules}
\acrodef{DM}{Data Mining}
\acrodef{MDLP}{Minimum Description Length Principle}
\acrodef{EDiRa}{Entropy-based Discretization for Ranking data}
\acrodef{RAC}{Ranking As Class}

\bibliographystyle{elsarticle-num}
\bibliography{labelranking}

\end{document}